\documentclass[acmtog]{acmart}
\AtBeginDocument{%
  }


\copyrightyear{2025}
\acmYear{2025}
\setcopyright{cc}
\setcctype{by}
\acmConference[SA Conference Papers '25]{SIGGRAPH Asia 2025 Conference Papers}{December 15--18, 2025}{Hong Kong, Hong Kong}
\acmBooktitle{SIGGRAPH Asia 2025 Conference Papers (SA Conference Papers '25), December 15--18, 2025, Hong Kong, Hong Kong}\acmDOI{10.1145/3757377.3763917}
\acmISBN{979-8-4007-2137-3/2025/12}

\acmSubmissionID{papers\_1753}

\usepackage{graphicx}
\usepackage{subcaption}
\usepackage{epigraph}
\usepackage{cleveref}
\usepackage{placeins}
\usepackage{wrapfig}

\usepackage{algorithm}
\usepackage[noend]{algpseudocode}

\newcommand{\ie}{\textit{i.e.}}
\newcommand{\eg}{\textit{e.g.}}

\usepackage{xcolor}
\usepackage{xspace}

\makeatletter
\DeclareRobustCommand\onedot{\futurelet\@let@token\@onedot}
\def\@onedot{\ifx\@let@token.\else.\null\fi\xspace}

\usepackage{tikz}
\usetikzlibrary{positioning, arrows.meta}

\newcommand{\R}{{\mathbb{R}}}

\def\eg{\emph{e.g}\onedot} 
\def\ie{\emph{i.e}\onedot}

\newcommand{\ourModel}{OmnimatteZero\xspace}

\newcommand{\specialcell}[2][c]{%
  \begin{tabular}[#1]{@{}c@{}}#2\end{tabular}}

 \newcommand\bigforall{\mbox{\Large $\mathsurround0pt\forall$}}

\usepackage{pifont}
\newcommand{\cmark}{\ding{51}}%
\newcommand{\xmark}{\ding{55}}%

\hypersetup{
  colorlinks=true,
  urlcolor=blue
}

\begin{document}

\title{OmnimatteZero: Fast Training-free Omnimatte with Pre-trained Video Diffusion Models}


\author{Dvir Samuel}
\email{dvirsamuel@gmail.com}
\affiliation{%
  \institution{Bar-Ilan University \& OriginAI}
  \country{Israel}
}

\author{Matan Levy}
\affiliation{%
  \institution{The Hebrew University of Jerusalem}
  \country{Israel}
}

\author{Nir Darshan}
\affiliation{%
  \institution{OriginAI}
  \country{Israel}
}

\author{Gal Chechik}
\affiliation{%
  \institution{Bar-Ilan University \& NVIDIA Research}
  \country{Israel}
}

\author{Rami Ben-Ari}
\affiliation{%
  \institution{OriginAI}
  \country{Israel}
}


\begin{teaserfigure}
  \centering
  \includegraphics[width=0.8\linewidth]{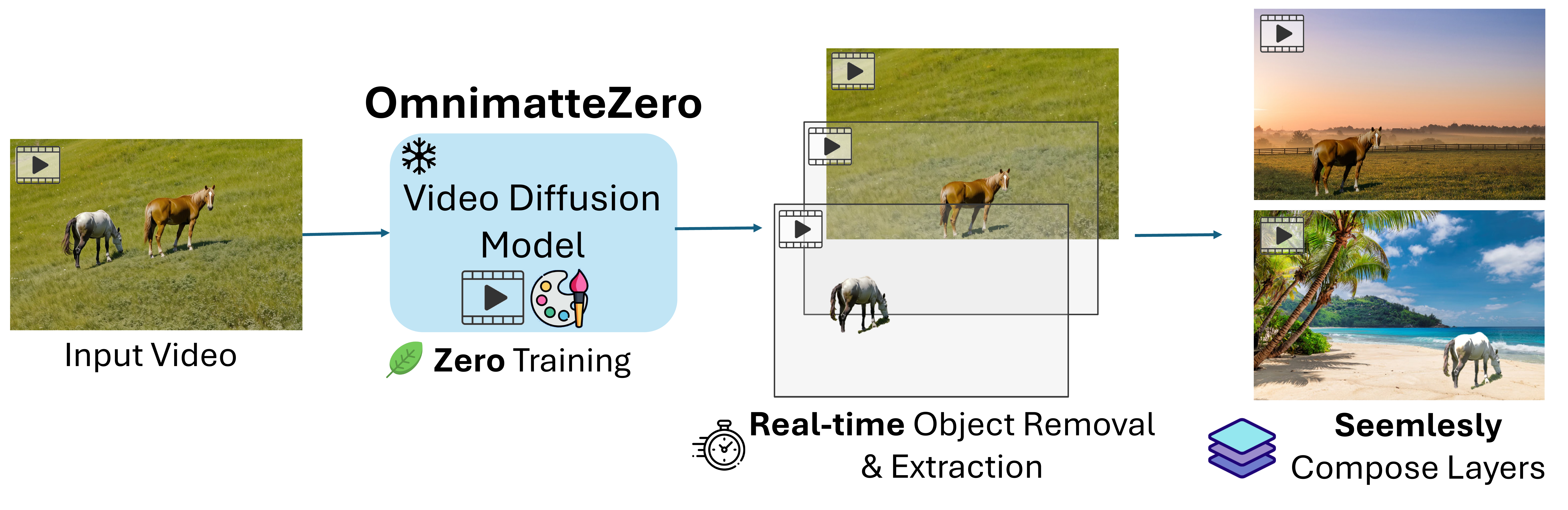}
    \caption{\textbf{OmnimatteZero} is the first training-free generative approach for Omnimatte, leveraging pre-trained video diffusion models to achieve object removal, extraction, and seamless layer compositions in just 0.04 sec/frame (on an A100 GPU). See video examples in the supplemental material.}
    \Description[<short description>]{<long description>}
    \label{fig:fig1}
\end{teaserfigure}

\begin{abstract}
In Omnimatte, one aims to decompose a given video into semantically meaningful layers, including the background and individual objects along with their associated effects, such as shadows and reflections. Existing methods often require extensive training or costly self-supervised optimization. In this paper, we present \textbf{OmnimatteZero}, a training-free approach that leverages off-the-shelf pre-trained video diffusion models for omnimatte. It can remove objects from videos, extract individual object layers along with their effects, and composite those objects onto new videos. 
These are accomplished by adapting zero-shot image inpainting techniques for video object removal, a task they fail to handle effectively out-of-the-box. To overcome this, we introduce temporal and spatial attention guidance modules that steer the diffusion process for accurate object removal and temporally consistent background reconstruction.  We further show that self-attention maps capture information about the object and its footprints and use them to inpaint the object's effects, leaving a clean background. Additionally, through simple latent arithmetic, object layers can be isolated and recombined seamlessly with new video layers to produce new videos. Evaluations show that OmnimatteZero not only achieves superior performance in terms of background reconstruction but also sets a new record for the fastest Omnimatte approach, achieving real-time performance with minimal frame runtime. 
\href{https://dvirsamuel.github.io/omnimattezero.github.io/}{\textcolor{blue}{Project Page.}}
\end{abstract}


\maketitle

\section{Introduction}
\label{sec:introduction}

Extracting or isolating objects in videos is a central problem in video understanding and editing, with important applications such as removing, modifying, or enhancing elements within a video scene. One important flavor of this task is video matting~\cite{chuang2002video}, which involves decomposing a video into a background layer and one or more foreground layers, each representing an individual object.
A particularly challenging form of video matting is \textit{Omnimatte}~\cite{representing_moving_1994, omnimatte}, where each foreground object is not only isolated but also includes associated effects such as shadows, reflections, and scene perturbations.
The layers extracted by video matting facilitate various video editing applications, including object removal, background replacement, retiming \cite{retimingVideo2020}, and object duplication \cite{generative-omnimatte}, many of which require estimating occluded background regions using techniques like video inpainting or omnimatting.  
%
%
%
%
%
%

However, despite their promise, existing methods for video inpainting and Omnimatte face significant limitations. Recent inpainting approaches~\cite{diffuEraser, propainter}, while trained on annotated datasets for object removal, fail to account for the nuanced effects that objects impart on the scene. Similarly, current Omnimatte methods demand computationally intensive self-supervised optimization for each video~\cite{omnimatte, omnimatte3D, omnimatteRF, generative-omnimatte}. Some approaches further rely on large hand-crafted training datasets~\cite{objectdrop, generative-omnimatte} or require 3D scene modeling using meshes or NeRFs~\cite{omnimatte3D, omnimatteRF}, resulting in hours of training and slow per-frame rendering.

Given the success of fast, training-free approaches in image editing~\cite{repaint, corneanu2024latentpaint}, a natural question arises: can these methods be effectively adapted to video editing? 

In this paper, we present a new approach called \textbf{OmnimatteZero}, the first training-free generative method to Omnimatte. \ourModel (named for its \textbf{zero} training and optimization requirements) leverages off-the-shelf pre-trained video diffusion models to remove all objects from a video, extract individual object layers along with their associated effects, and finally composite them onto new backgrounds. This entire process is performed efficiently during diffusion inference, making it significantly faster than current methods. See Figure \ref{fig:fig1} for illustration.

We begin by adapting training-free image inpainting methods~\cite{repaint,SDEdit} to handle object inpainting in videos using diffusion models. However, we find that simply applying these frame-by-frame techniques to videos leads to poor results: missing regions are not filled convincingly, and temporal consistency breaks down.
To address this, we propose two attention guidance modules that can be easily integrated into pre-trained video diffusion models. These modules guide the diffusion process to achieve both accurate object removal and consistent background reconstruction.
The first module, \textbf{Temporal Attention Guidance (TAG)}, uses background patches from nearby frames. By modifying self-attention layers, it encourages the model to use these patches to reconstruct masked regions, improving temporal coherence. The second module, \textbf{Spatial Attention Guidance (SAG)}, operates within each frame to refine local details and enhance visual plausibility.

We also observe that self-attention maps contain cues not just about the object but also its indirect effects, such as shadows and reflections. By leveraging this information, our method can inpaint these subtle traces, yielding cleaner backgrounds.
For foreground layer extraction, we find that the object (and its traces) can be isolated by subtracting the latent representation of the clean background from the original video latent. This extracted object layer can then be seamlessly inserted into new backgrounds by simply adding the latents, enabling fast and flexible video composition.

To validate our approach, we apply it to two video diffusion models, LTXVideo \cite{LTXVideo} and Wan2.1 \cite{wan2.1}, and evaluate it on standard Omnimatte benchmarks both quantitatively and qualitatively. Our method outperforms all supervised and self-supervised existing inpainting and Omnimatte techniques across all benchmarks in terms of background reconstruction and object layer extraction. Notably, OmnimatteZero requires no training or additional optimization, functioning entirely within the denoising process of diffusion models. This positions OmnimatteZero as the fastest Omnimatte technique to date.

\section{Related Work}
\label{sec:related_work}

\subsection{Omnimatte}

The first Omnimatte approaches~\cite{kasten2021layered, omnimatte} use motion cues to decompose a video into RGBA matte layers, assuming a static background with planar homographies. Optimization isolates dynamic foreground layers, taking 3 to 8.5 hours per video and  $\sim$2.5 seconds per frame to render.
Later methods enhanced Omnimatte with deep image priors~\cite{omnimattesp, factormatte} and 3D representations~\cite{omnimatteRF, omnimatte3D, d2nerf}, but still rely on strict motion assumptions and require extensive training ($\sim$6 hours) with rendering times of 2.5-3.5 seconds per frame.
A recent generative approach~\cite{generative-omnimatte} uses video diffusion priors to remove objects and effects, with test-time optimization for frame reconstruction, taking $\sim$9 seconds per frame.
In this paper, we propose a novel, training and optimization-free method using off-the-shelf video diffusion models, achieving real-time performance in 0.04 seconds per frame.

\subsection{Video Inpainting}
Earlier video inpainting methods~\cite{3dgated_conv, chuan_video_inpainting, yuan_video_completion} used 3D CNNs for spatiotemporal modeling but struggled with long-range propagation due to limited receptive fields. Pixel-flow approaches~\cite{chen_flow_video_completion, kaidong_video_inp, kaidong_inertia_guided} improved texture and detail restoration by utilizing adjacent frame information. Recently, \citet{propainter} introduced ProPainter, which enhances reconstruction accuracy and efficiency by leveraging optical flow-based propagation and attention maps. VideoPainter \cite{videopainter} proposed a dual-stream video inpainting method that injects video context into a pre-trained video diffusion model using a lightweight encoder. \cite{diffuEraser} proposed DiffuEraser, which combines a pre-trained text-to-image diffusion model with BrushNet, a feature extraction module, to generate temporally consistent videos. 
In this paper, we propose a single-step object inpainting approach, directly applied during video diffusion model inference, implicitly leveraging spatial and temporal information from adjacent frames for inpainting. 

\subsection{Diffusion Transformers for Video Generation}
Recent advancements in diffusion models have significantly improved text-to-video generation. These models, trained on large datasets of video-text pairs, can generate visually compelling videos from text descriptions. 
Recent work introduced LTX-Video \cite{LTXVideo}, a real-time transformer-based latent diffusion model capable of generating videos at over 24 frames per second. It builds on 3D Video-VAE with a spatiotemporal latent space, a concept also used in Wan2.1 \cite{wan2.1} and the larger HunyuanVideo \cite{kong2024hunyuanvideo}. Unlike traditional methods that treat the video VAE and denoising transformer separately, LTX-Video integrates them within a highly compressed latent space (1:192 compression ratio), optimizing their interaction. This spatiotemporal downscaling results in $32 \times 32 \times 8$ pixels per token while maintaining high-quality video generation. Similarly, HunyuanVideo employs a large-scale video generation framework, integrating an optimized VAE and diffusion-based transformer to achieve excellent visual quality and motion consistency. These models demonstrate the trend of combining high-compression VAEs with transformer-based diffusion models for efficient and scalable video generation.

\begin{figure}[t!]
	\centering
	\includegraphics[width=\linewidth]{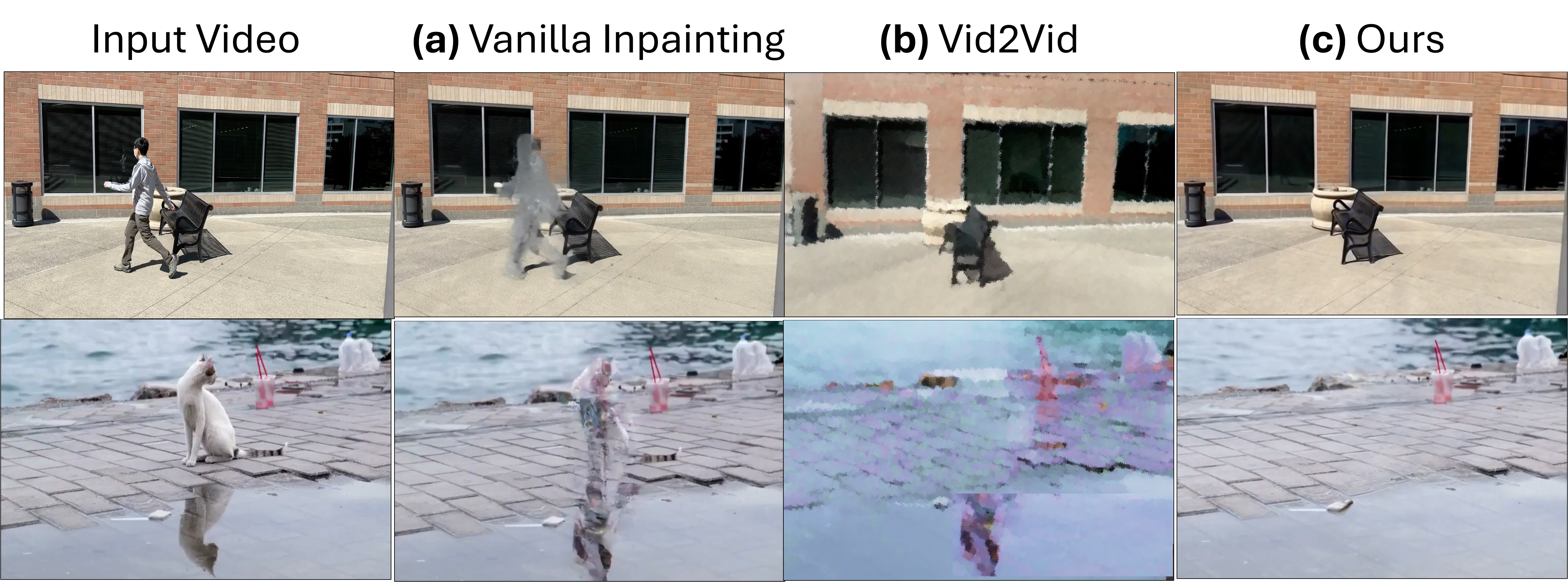}
    \Description[motivation figure]{motivation figure}
	\caption{Comparison of object removal results using \textbf{(a)} vanilla image inpainting extended to video, \textbf{(b)} Vid2Vid zero-shot inpainting, and \textbf{(c)} our guidance-based approach. Vanilla methods fail to maintain temporal consistency or clean background reconstruction, while our method achieves coherent, artifact-free inpainting across frames.}
	\label{fig:motivation}
\end{figure}

\begin{figure*}[t!]
	\centering
	\includegraphics[width=0.9\linewidth]{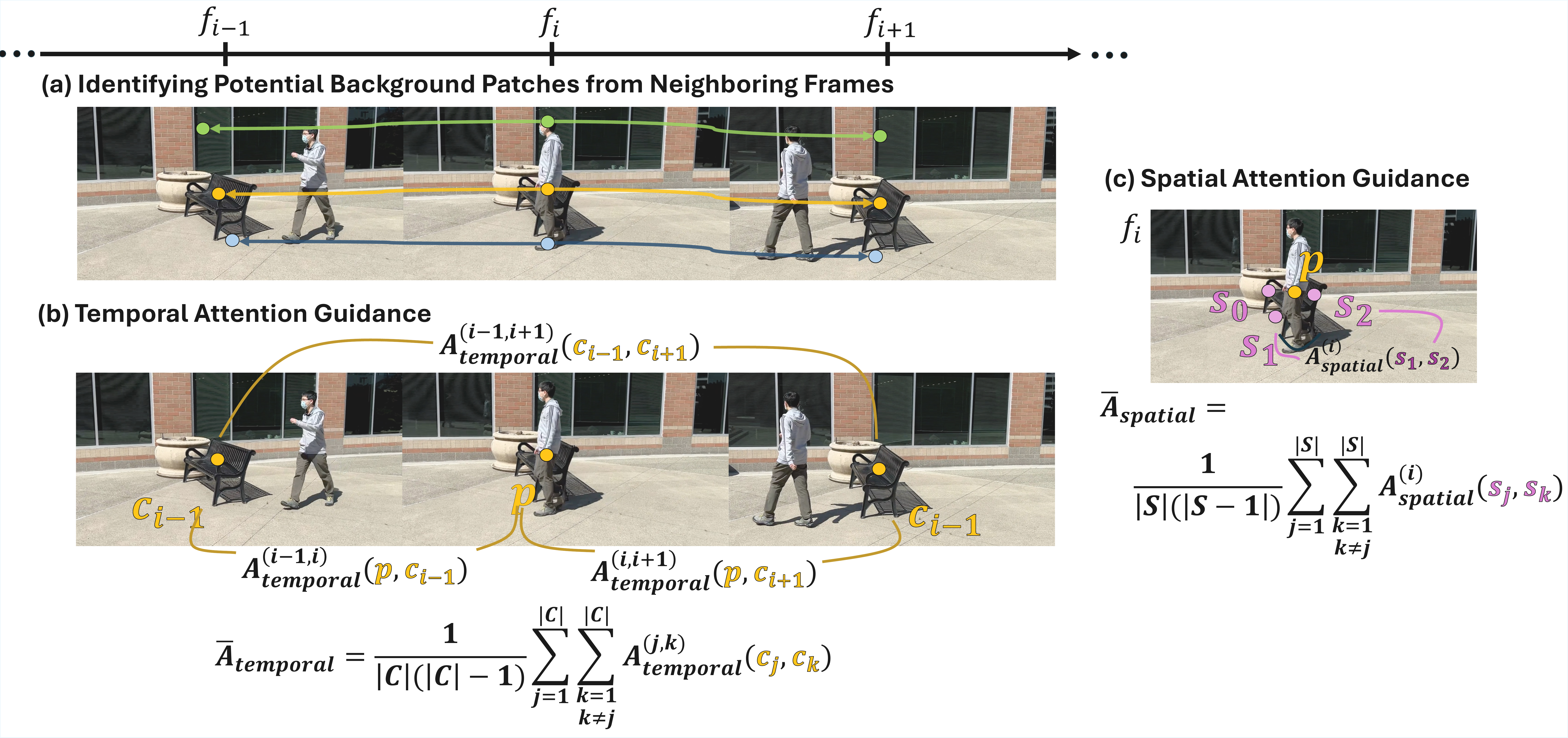}
    \Description[method figure]{method figure}
	\caption{\textbf{Overview of our Object Removal strategy in OmnimatteZero.}
\textbf{(a)} We first identify potential background correspondences across frames. 
\textbf{(b)} Temporal Attention Guidance (TAG): Temporal attention scores between a foreground point and its background correspondences are replaced with the average attention between all background pairs, promoting consistent inpainting across time.
\textbf{(c)} Spatial Attention Guidance (SAG): Within a frame, the attention from a foreground point to nearby background points is adjusted to reflect the mean attention among background points themselves, improving inpainting quality when temporal context is unavailable.}
	\label{fig:method}
\end{figure*}

\section{Motivation: The failure of image inpainting approaches for videos}
\label{sec:motivation}

We begin by extending training-free image-diffusion inpainting techniques to video diffusion models. Our reference point is \emph{vanilla image inpainting}~\cite{repaint}, a straightforward method that starts with a random noise at the masked area and, at each diffusion step $t$, injects Gaussian noise $\epsilon_{t}$ into the original background area so the denoiser can reconstruct the masked region while preserving the background during denoising.  For video, we apply it across the full video latent to maintain both spatial and temporal coherence, since frame-by-frame inpainting leads to temporal artifacts like flickering.
In practice, this vanilla approach does preserve the original background alignment, however causes the masked areas to degrade into incoherent, temporally unstable blobs (Fig.~\ref{fig:motivation}a). We attribute this breakdown to a core difference between image and video models: image inpainting ensures spatial consistency within a single frame, while video inpainting must also maintain temporal consistency across frames. Unlike image inpainting, which relies heavily on hallucination, video inpainting must infer missing content from both spatial and temporal context, making it significantly more challenging.
We further examine a second zero-shot inpainting method: \emph{Vid2Vid} \cite{LTXVideo} (an adaptation from Img2Img~\cite{SDEdit}). This method applies heavy noise inside the mask and lighter noise elsewhere in the latent video, then denoises the entire latent. Vid2Vid effectively removes the target object and synthesizes plausible fills, however, it also introduces artifacts into the background and motion (Fig.\ref{fig:motivation}b). 

While recent diffusion-based solutions \cite{videopainter, generative-omnimatte} overcome these issues by training dedicated video diffusion models for object removal, we pose the following question: \emph{Can we develop a training free method that simultaneously (i) convincingly fills the masked region, (ii) preserves the background so that it is indistinguishable from the original video, and (iii) eliminates all residual evidence of the object’s presence, including secondary effects like shadows and reflections?}. 

\section{Preliminaries: Spatio–Temporal Attention in Video Diffusion Transformers}

Video diffusion transformers interleave \textbf{spatio}-\textbf{temporal} self-attention modules in each diffusion block to model both fine-grained structure within frames and consistent dynamics across time.
Given a video, it's latent tensor is $Z \in \mathbb{R}^{f \times w \times h \times c}$, where $f$ is the number of temporally compressed frames, and $w$, $h$, and $c$ are the width, height, and channel dimensions in latent space. We reshape $Z$ into $Z \in \mathbb{R}^{(f \times n) \times c}$, where $n = h \times w$ is the number of tokens per frame.
Self‐attention is then applied over all $f \times n$ tokens:
$$
A \;=\;\mathrm{Attention}\bigl(Q(Z),\,K(Z)\bigr)
\;\in\; \mathbb{R}^{(f \times n) \times (f \times n)}.
$$ where $Q,K \in \mathbb{R}^{w \times h \times c}$ 
are the queries and keys, respectively.
This matrix $A$ encodes both spatial and temporal interactions in one operation.
We denote tokens by $(i,p)$ with $i\!=\!1,\dots,f$ the frame index and $p\!=\!1,\dots,n$ the spatial position.
\emph{Spatial attention} within frame $i$ (i.e.\ interactions among tokens $(i,p)$ and $(i,q)$) is given by the in-frame weights
$      A_{\mathrm{spatial}}^{(i)}(p,q)
      \;=\;A_{(i,p),(i,q)}.$
\emph{Temporal attention} between frame $i$ to frame $j\neq i$ (i.e.\ interactions between tokens $(i,p)$ and $(j,q)$) is given by 
$A_{\mathrm{temporal}}^{(i,j)}(p,q)\;=\;A_{(i,p),(j,q)}.$



\section{Method: OmnimatteZero}

In this section, we first describe how given a video  $\mathcal{V} \in \mathbb{R}^{F \times W \times H \times 3}$ and an object mask for each frame $M_{obj} \in \{0,1\}^{F \times H \times W}$,  one can use off-the-shelf video diffusion models to remove the target object (Section \ref{subsec:object_removal}) and its associated visual effects (Section \ref{subsec:effects}). Next, we detail our strategy for isolating foreground objects together with their residual effects (Section \ref{subsec:foreground}). Finally, we demonstrate how to recombine these object–effect layers onto arbitrary background videos (Section \ref{subsec:composition}), enabling flexible and realistic video editing.

In summary, our pipeline expands masks to capture object effects, inpaints them for clean 
backgrounds, isolates foregrounds via latent arithmetic, and seamlessly recomposes them 
onto new scenes.

\subsection{Object Removal}
\label{subsec:object_removal}
Zero-shot video inpainting with diffusion models ~\cite{SDEdit, repaint} frequently struggles because the model must \emph{hallucinate} content that is coherent both spatially and temporally ( \Cref{sec:motivation}). Recent studies show that the self-attention mechanism within diffusion models is crucial for maintaining coherence during model generation~\cite{enhance}. However, it remains unclear how self-attention can effectively guide the inpainting process.

Our core observation is that in real-world videos, the necessary background information is often readily available in neighboring frames. This principle has been leveraged in traditional video editing techniques \cite{classical1, classical2}. However, they often produce non-realistic results or do not scale for long videos. Motivated by this, we introduce a method that explicitly guides a video diffusion model to reconstruct missing pixels by leveraging spatial and temporal context, rather than relying on the model to infer this implicitly. To further enhance spatial inpainting within individual frames, particularly when background details are absent from adjacent frames (e.g., due to static scenes or lack of camera/object motion), we exploit contextual cues from regions surrounding the masked area within the same frame. These temporal and spatial guidance are facilitated through careful manipulations of spatial and temporal self-attention. 

We thus introduce two novel modules: \emph{Temporal Attention guidance} (TAG) and \emph{Spatial Attention Guidance} (SAG). These modules seamlessly integrate into existing pre-trained video diffusion models, effectively steering the diffusion process toward object removal and coherent background reconstruction.
\subsubsection*{\textbf{Identifying Potential Background Patches from Neighboring Frames}}
To accurately inpaint masked areas in a specific frame, it is essential to identify corresponding background patches from neighboring frames, while preserving awareness of the video's underlying 3D structure. This task can be effectively addressed using Track-Any-Point (TAP-Net) \cite{doersch2023tapir}, a real-time tracking framework. TAP-Net captures spatial and temporal coherence by tracking points from one frame to their corresponding locations across subsequent frames, as illustrated in \Cref{fig:method}a, where points from frame $f_{i}$ are matched across other frames.
Formally, given a point $p$ on the object in frame $i$, TAP-Net identifies its corresponding point set among frames $C = \{C_{j} | j \ne i\}$. To ensure that only background regions are considered, we discard points in $C$ that intersect the object mask $\mathcal{M}$ in their respective frames.

\subsubsection*{\textbf{Temporal Attention Guidance (TAG)}}
To effectively guide the denoising process toward inpainting using neighboring frames, we explicitly set the temporal attention score between a foreground point $p$ (in frame $i$) and each of its background correspondences $c \in C$ to be the mean temporal attention observed between all distinct pairs of corresponding background points (see Figure~\ref{fig:method}b). Formally, we first compute the mean temporal attention

\begin{equation}
    \bar{A}_{temporal}=\frac{1}{|C|(|C|-1)}{\sum_{j =1}^{|C|}{\sum_{\substack{k=1 \\ k \ne j}}^{|C|}{A^{(j,k)}_{temporal}(c_j,c_k)}}}.
    \label{eq:mean_bg_attention1}
\end{equation}

Since the attention is bidirectional, the denominator $|C|(|C|-1)$ correctly enumerates all ordered pairs $(c_j, c_k)$ where $j \ne k$, ensuring that each directional interaction is counted exactly once.

We then replace the temporal-attention score from $p$ to every background point $c$ as follows:
$$A_{temporal}^{(i,j)}(p,c_j) = \bar{A}_{temporal}, \quad \bigforall c_j \in C.$$

This assignment transfers a consensus cue from the entire background set to the foreground point, encouraging consistent inpainting across frames.

\subsubsection*{\textbf{Spatial Attention Guidance (SAG)}}
To further improve spatial inpainting within each frame and to address cases where a point lacks correspondences in other frames, we reinforce spatial attention within the same frame. Specifically, for any foreground point $p$ in frame $i$, we set its spatial attention scores with \textit{surrounding} background points $S$ to be the mean spatial attention computed among all pairs of background points within the same frame (see Figure \ref{fig:method}c). Formally:

\begin{equation}
    \bar{A}_{spatial}=\frac{1}{|S|(|S|-1)}{\sum_{j =1}^{|S|}{\sum_{\substack{k=1 \\ k \ne j}}^{|S|}{A^{(I)}_{spatial}(s_j,s_k)}}}.
    \label{eq:mean_bg_attention}
\end{equation}

Finally, we replace the spatial-attention score from $p$ to every background point $s$ as follows:
$$A^{i}_{\text{spatial}}(p, s) = \bar{A}_{spatial}, \quad \bigforall s \in S.
$$
This strategy ensures coherent spatial inpainting even in the absence of temporal information.


\begin{figure}[t!]
	\centering
	\includegraphics[width=0.8\linewidth]{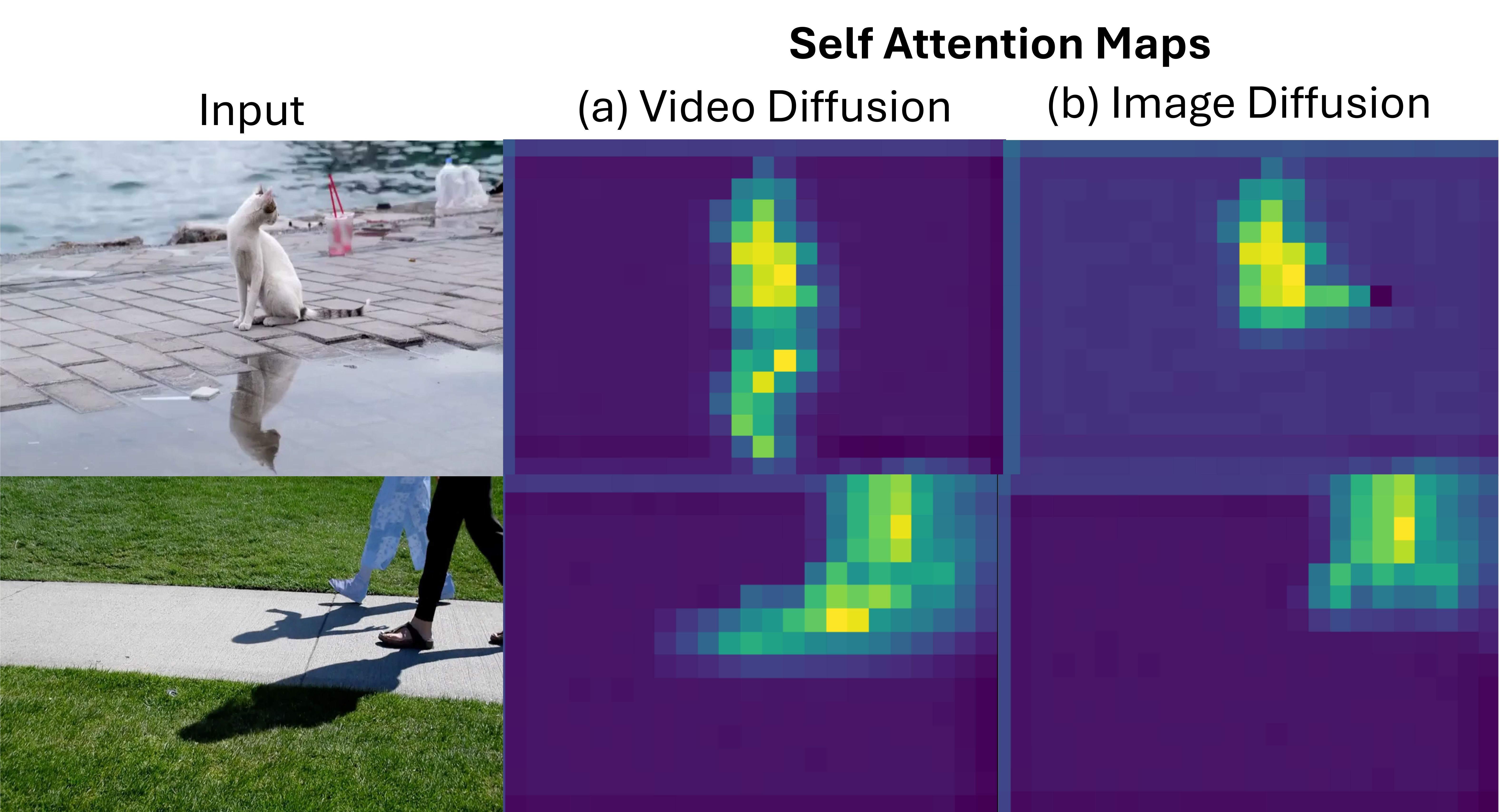}
    \Description[Description]{Description}
	\caption{\textbf{Self-attention maps} from \textbf{(a)} LTX Video diffusion model and \textbf{(b)} Stable Diffusion (image based).  The spatio-temporal video latent "attends to object associated effects" (e.g., shadow, reflection) where, image models struggles to capture these associations.}
	\label{fig:attention}
\end{figure}

\subsection{Removing Associated Object Effects}
\label{subsec:effects}

We are interested in masking the object with its associated effects.
Recently, \cite{generative-omnimatte} showed that pretrained video diffusion models can associate objects with their effects by analyzing self-attention maps between query and key tokens related to the object of interest. They leveraged this insight to train diffusion models for object removal along with its associated footprint. 
In contrast to \cite{generative-omnimatte}, we propose to directly derive the masks from the attention maps, allowing a training-free object removal approach. Specifically, we apply a single noising and denoising step on the video latent and compute the attention weights between query and key tokens associated with the object. 
More precisely, for each latent frame $i$ at diffusion layer $l$, we calculate the attention map as follows:
\begin{equation}
A_{i}^{l} = \text{softmax}\left(\frac{Q_{i}^{l} (M_{obj} \odot K_{i}^{l})^{T}}{\sqrt{c}}\right) \in \R^{N \times N}
\end{equation}
where $Q^{l}_{i}, K^{l}_{i} \in \R^{w \times h \times c}$ are the queries and keys respectively at layer $l$ and frame $i$, $N=w \times h$ and $c$ is the number of channels of the latent.

We then compute a soft-mask per-frame, $\mathcal{M}_i \in [0,1]^{w \times h \times c}$, which is aggregated across all diffusion layers: 
\begin{equation}
\mathcal{M}_{i} = \frac{1}{L}\sum_{l=1}^{L} A_{i}^{l}
\end{equation}
where $L$ is the number of layers in the video diffusion model. 
Then we extend $\mathcal{M}_i$ to the whole video by concatenating all frame masks channel-wise $\mathcal{M}_{obj} = \text{concat}\{\mathcal
{M}_1, ... \mathcal{M}_f \} \in \R^{w \times h \times f}$ deriving a soft-mask latent. To obtain a binary mask we perform Otsu thresholding \cite{otsu} for each latent frame, $\hat{M}_{obj} = \mathbb{I} (\mathcal{M}_{obj})$. This new mask replaces the mask $M_{obj}$ provided as input by the user. 
Figure \ref{fig:attention}(a) shows self-attention maps from LTX-Video \cite{LTXVideo} of two video frames. The attention maps reveal the model’s ability to localize not only primary objects—like the cat and the walking person—but also their associated physical effects, such as reflections and shadows. This demonstrates the model’s robustness in consistently tracking these cues even when similar visual patterns are present elsewhere in the scene.

Interestingly, to the best of our knowledge, this approach has not been explored for masking object effects (e.g shadows) in images. Unlike video diffusion models, image models do not capture object effects from still images \cite{objectdrop}. \Cref{fig:attention}(b) illustrates this by showing self-attention maps extracted using StableDiffusion~\cite{StableDiffusion}, a text-to-image diffusion model, which demonstrates that the object does not attend to its associated effects.
We believe that this aligns with the principle of \textit{common fate} in Gestalt psychology \cite{gestaltKohler1992}, which suggests that elements moving together are perceived as part of the same object. Video diffusion models seem to implicitly leverage this principle by grouping objects with their effects through motion.

\begin{figure}[t!]
	\centering
	\includegraphics[width=1\linewidth]{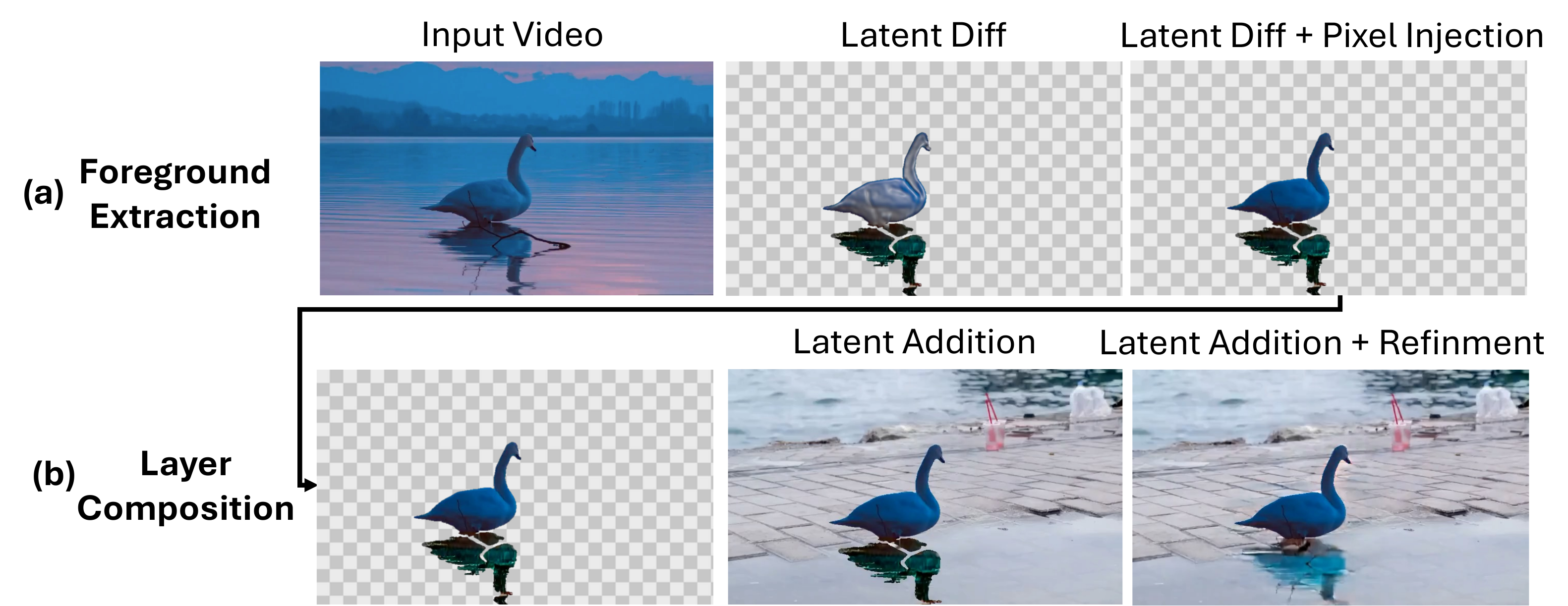}
    \Description[Description]{Description}
	\caption{\textbf{(a) Foreground Extraction:} The target object is extracted by latent code arithmetic, subtracting the background video encoding from the object+background latent (Latent Diff). This initially results in distortions, which are later corrected by replacing pixel values using the original video and a user-provided mask (Latent Diff + Pixel injection).
\textbf{(b) Layer Composition:} The extracted object layer is added to a new background latent (Latent Addition). To improve blending, a few steps of noising-denoising are applied, yielding a more natural integration of the object into the new scene (Latent Addition + Refinement). See video examples in the supp material.}
	\label{fig:extraction_composition}
\end{figure}

\begin{table*}[t!]
\centering
\caption{\textbf{Quantitative comparison: Background Reconstruction. } \textbf{OmnimatteZero} outperforms all omnimatte and video inpainting methods, achieving the best PSNR and LPIPS without training or per-video optimization. It also runs significantly faster, with OmnimatteZero [LTXVideo] at 0.04s per frame. "-" denotes missing values due to unreported data or unavailable public code.}
\scalebox{0.8}{
\begin{tabular}{l|cccccc|ccc|cc}
\toprule
Scene & \multicolumn{3}{c}{Movie} & \multicolumn{3}{c|}{Kubric} & \multicolumn{3}{c}{Average} & &\\
Metric & PSNR$\uparrow$ & LPIPS$\downarrow$ & SSIM$\uparrow$ & PSNR$\uparrow$ & LPIPS$\downarrow$ & SSIM$\uparrow$ & PSNR$\uparrow$ & LPIPS$\downarrow$ & SSIM$\uparrow$& \specialcell{$T_{train}$\\(hours)} & \specialcell{$T_{run}$\\(s/frame)}\\ \midrule
ObjectDrop~\cite{objectdrop} & 28.05 & 0.124 & -& 34.22 & 0.083 & - & 31.14 & 0.104 & -& - & -\\
\midrule
Video Repaint [LTXVideo]*~\cite{repaint} & 20.13 & 0.252 & -& 21.15 & 0.289 & -& 20.64 & 0.271 & -& \textbf{0} & 0.4\\
Video Repaint [Wan2.1]*~\cite{repaint} & 21.44 &  0.244 & -& 24.16 & 0.261 & -& 22.8 & 0.253 & -& \textbf{0} & 32\\
Temporal Enhance [LTXVideo]*~\cite{enhance} & 21.33 & 0.248 & -& 23.01 & 0.281 & -& 22.8 & 0.265 & -& \textbf{0} & \textbf{0.04}\\
Temporal Enhance [Wan2.1]*~\cite{enhance} & 21.93 & 0.222 & -& 24.26 & 0.253 & -& 23.01 & 0.237 & -& \textbf{0} & 3.2\\
Lumiere inpainting~\cite{lumiere} & 26.62 & 0.148 & -& 31.46 & 0.157 & -& 29.04 & 0.153 & -& - & 9\\
Propainter~\cite{propainter} & 27.44 & 0.114 & -& 34.67 & 0.056 & -& 31.06 & 0.085 & -& - & 0.083\\
DiffuEraser~\cite{propainter} & 29.51 & 0.105 & -& 35.19 & 0.048 & -& 32.35 & 0.077 & -& - & 0.8\\
\midrule
Ominmatte~\cite{omnimatte} & 21.76 & 0.239 & 0.736& 26.81 & 0.207 & 0.831 & 24.29 & 0.223 & 0.783 & 3 & 2.5\\
D2NeRF~\cite{d2nerf} & - & - & -& 34.99 & 0.113 & 0.887 & - & - & - & 3 & 2.2\\
LNA~\cite{kasten2021layered} & 23.10 & 0.129 & 0.847 & - & - & - & - & - & -& 8.5 & 0.4\\
OmnimatteRF~\cite{omnimatteRF} & 33.86 & 0.017 & 0.981 & 40.91 & 0.028 & 0.970 & 37.38 & 0.023 & 0.975& 6 & 3.5\\
Generative Omnimatte \cite{generative-omnimatte} & 32.69 & 0.030 & 0.989 & 44.07 & 0.010 & 0.981& 38.38 & 0.020 & 0.985 & - & 9\\ 
\midrule
\textbf{OmnimatteZero [LTXVideo]} (\textbf{Ours})  & \textbf{35.11} & \textbf{0.014} & \textbf{0.992} & \textbf{44.97} & 0.010 & \textbf{0.988} & \textbf{40.04} & \textbf{0.012} & \textbf{0.990} & \textbf{0} & \textbf{0.04}\\ 
\textbf{OmnimatteZero [Wan2.1]} (\textbf{Ours}) & \textbf{34.12} & 0.017 & \textbf{0.993} & \textbf{45.55} & \textbf{0.006} & \textbf{0.987} & \textbf{39.84} & \textbf{0.011} & \textbf{0.990} & \textbf{0} & 3.2\\ 
\bottomrule
\end{tabular}
}

\label{tab:quantitative_comparison}
\end{table*}

\subsection{Foreground Extraction}
\label{subsec:foreground}
We can now use our object removal approach to extract object layers along with their associated effects. To isolate a specific object, we apply our method twice: first, removing all objects from the video, leaving only the background, denoted as $ V_{bg} \in \mathbb{R}^{F \times W \times H \times 3} $ with its corresponding latent $ Z_{bg} $; second, removing all objects except the target object, resulting in a video of the object with the background, denoted as $ V_{obj+bg} \in \mathbb{R}^{F \times W \times H \times 3} $ with its corresponding latent $ Z_{obj+bg} $.
 We can now derive the video of the object isolated from the background by simply \textbf{subtracting} the two latents $Z_{obj} = Z_{obj+bg} - Z_{bg}$. Applying thresholding on $Z_{obj}$  results in a latent that is decoded to a video $V_{obj}$ containing only the object and its associated effects (see \Cref{fig:extraction_composition}a Latent Diff). While the extracted effects appear convincing, the object itself often suffers from distortions due to latent values falling outside the diffusion model's learned distribution. To address this issue, we make use of the information available in the pixel-space. We refine the object’s appearance by replacing its values in the pixel-space with those from the original video, based on the user provided mask:
\begin{equation}
V_{obj} = M^p_{obj} \odot V_{obj+bg} + \left( 1-M^p_{obj} \right) \odot V_{obj}.   
\end{equation}

This correction preserves the object's fidelity while maintaining its associated effects, resulting in high-quality object layers (see \Cref{fig:extraction_composition}a Latent Diff + Pixel Injection). A comparison of OmnimatteZero with other baselines for foreground extraction is provided in the Supplemental.

\textit{Alpha-Channel Extraction:}
The alpha-channel is obtained from the soft mask $M_i$ (Eq.~4) by decoding it with 
the video VAE using only the spatial and temporal upsampling layers, skipping 
feature re-encoding. This ensures (1) \emph{softness}, the output remains a soft mask 
that captures graded transitions around boundaries and effects, and (2) \emph{alignment},
the decoded alpha-channel is upsampled to the original resolution, maintaining 
pixel-wise correspondence with the RGB frames. The result is a temporally and 
spatially consistent matte that preserves soft boundaries, shadows, and reflections, 
enabling realistic compositing and evaluation.

\subsection{Layer Composition}
\label{subsec:composition}
With the object layers extracted, we can now seamlessly compose them onto new videos by \textbf{adding} the object layer latent to a {\it new} background latent $Z^{N}_{bg}$. Specifically,
\begin{equation}
Z^{N}_{obj+bg} = Z^{N}_{bg} + Z_{obj}.
\end{equation}

Figure \ref{fig:extraction_composition}b (Latent Addition) illustrates the initial result with some residual inconsistencies, which we finally fix, by applying a 3 noising-denoising steps. This process helps smooth transitions between the video background and foreground layers, resulting in a more natural and cohesive video (Figure \ref{fig:extraction_composition}b Latent Addition + Refinment).
\section{Experiments}
\label{sec:results}
Following \cite{generative-omnimatte}, we evaluate \ourModel on three applications: \textbf{(1) Background Reconstruction}, where the foreground is removed to recover a clean background; \textbf{(2) Foreground Extraction}, where objects are extracted, together with their associated effects (\eg shadows and reflections); and \textbf{(3) Layer Composition}, where extracted elements are reinserted into new backgrounds while preserving visual consistency. For qualitative results, each figure shows one representative frame per video; full videos are in the supplementary material.




\begin{figure*}[t!]
	\centering
	\includegraphics[width=\linewidth]{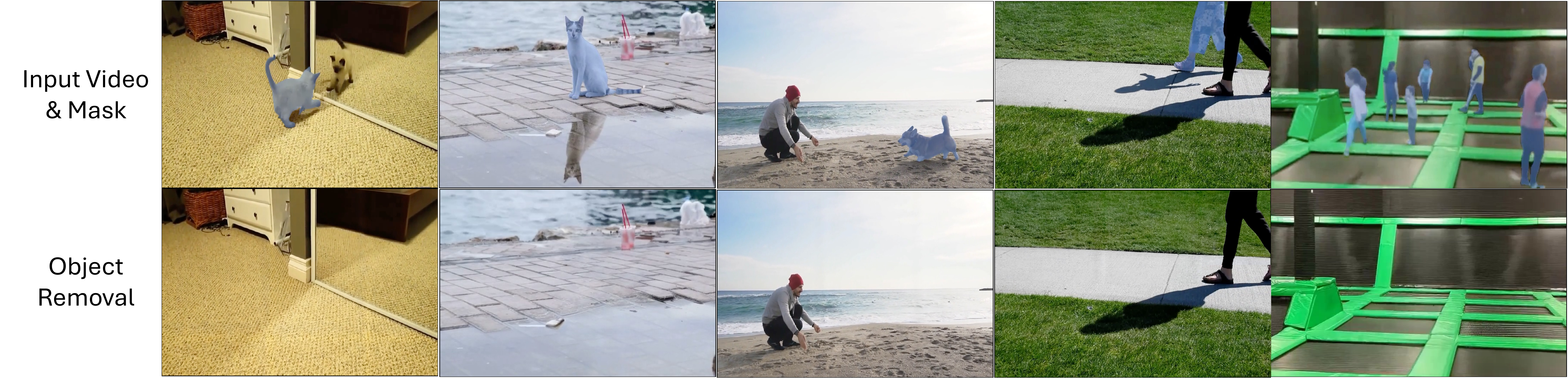}
    \Description[Description]{Description}
	\caption{\textbf{Qualitative Results: Object removal and background reconstruction.} The first row shows input video frames with object masks, while the second row presents the reconstructed backgrounds. Our approach effectively removes objects while preserving fine details, reflections, and textures, demonstrating robustness across diverse scenes. Notice the removal of the cat’s reflection in the mirror and water, the shadow of the dog and bicycle (with the rider), and the bending of the trampoline when removing the jumpers. See video examples in the supplemental material 
    }
	\label{fig:qualitative_results}
\end{figure*}


\subsection{Background Layer Reconstruction}

\paragraph{Compared methods:} We conducted a comparative evaluation of our approach with several SoTA methods. These methods fall into four categories: \textbf{(A) Omnimatte methods} that are trained to decompose a given video into layers: Omnimatte~\cite{omnimatte}, D2NeRF~\cite{d2nerf}, LNA~\cite{kasten2021layered}, OmnimatteRF~\cite{omnimatteRF} and Generative Omnimatte~\cite{generative-omnimatte}. \textbf{(B) Video inpainting methods} that are trained to remove objects from a video: RePaint~\cite{repaint} adapted for video, Temporal Enhance~\cite{enhance} applied to vanilla inpainting (Sec \ref{fig:motivation}) which enhances general temporal attention during denoising, Lumiere inpainting~\cite{lumiere}, Propainter~\cite{propainter}, DiffuEraser~\cite{diffuEraser}, and VideoPainter~\cite{videopainter}. Finally,  \textbf{(C) An image inpainting method} that is applied for each video frame independently: ObjectDrop~\cite{objectdrop}.  

\indent\textit{Datasets and Metrics:} We evaluate our method on two standard Omnimatte datasets: Movies~\cite{omnimatteRF} and Kubric~\cite{d2nerf}. These datasets provide object masks for each frame and ground truth background layers for evaluation. All methods are assessed using PSNR, SSIM, and LPIPS metrics to measure the accuracy of background reconstruction, along with comparisons of training and runtime efficiency on a single A100 GPU. 

\indent\textit{Quantitative results:} \Cref{tab:quantitative_comparison} presents a comparison of \ourModel with the SoTA Omnimatte and inpainting methods on the Movies and Kubric benchmarks. It shows that OmnimatteZero outperforms existing supervised and self-supervised methods designed for object inpainting or omnimatte. It achieves the highest PSNR and lowest LPIPS on both the Movie and Kubric datasets, demonstrating superior background reconstruction. Specifically, OmnimatteZero [LTXVideo] achieves a PSNR of 35.11 (Movie) and 44.97 (Kubric), surpassing Generative Omnimatte~\cite{generative-omnimatte}, all while requiring \textbf{no training or per-video optimization}. Notably, this improvement is not due to a stronger video generation model, as both OmnimatteZero and Video RePaint~\cite{repaint} use the same generator, yet Video RePaint records the lowest PSNR and highest LPIPS across all benchmarks.

Our method is also significantly faster. OmnimatteZero [LTXVideo] runs at just 0.04s per frame (or 24 frames per second). These results establish OmnimatteZero as the first training-free, real-time video matting method, offering both superior performance and efficiency.

\begin{figure}[t!]
	\centering
\includegraphics[width=\linewidth]{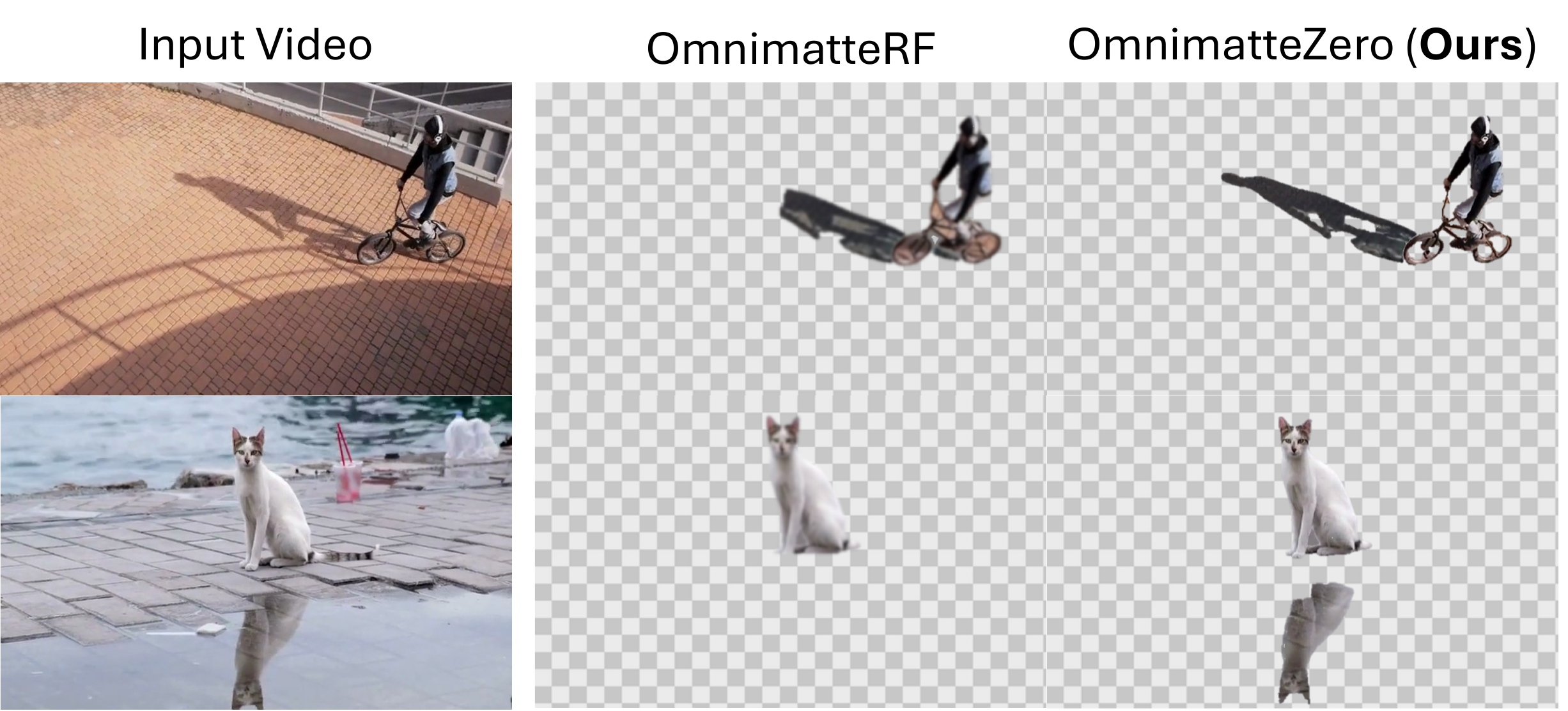}
\Description[Description]{Description}
	\caption{\textbf{Qualitative Comparison: Foreground Extraction.}  Foreground extraction comparison between \ourModel and OmnimatteRF \cite{omnimatteRF}. Our method accurately captures both the object and its associated effects, such as shadows and reflections, in contrast to OmnimatteRF, missing or distorting shadows (row 1) and reflections (row 2). }
	\label{fig:qualitative_foreground}
\end{figure}

\indent\textit{Qualitative results:} 
\Cref{fig:qualitative_results} presents qualitative results of our object removal method across various scenes. The first row shows input video frames with masked objects, while the second row displays the reconstructed backgrounds. The videos include (1) a cat running toward a mirror, (2) a static cat looking left and right, (3) a dog running toward a man, (4) two people walking, and (5) multiple people jumping on a trampoline.
Our method effectively removes objects like people and animals even when similar objects and effects appear in the video, while preserving fine details and textures. Notably, in the first two columns, \ourModel eliminates the cat without leaving mirror or water reflections. The last column further demonstrates its ability to remove objects while maintaining scene integrity, even correcting the trampoline’s bending effect after removing the jumper.
\Cref{fig:qualitative_1} \ref{fig:qualitative_2} presents a qualitative comparison of \textbf{OmnimatteZero} with SoTA omnimatte and video inpainting methods. Our method consistently produces cleaner and more temporally coherent background reconstructions with fewer artifacts and distortions, especially in challenging scenes involving motion, shadows, and texture continuity.
The figures includes diverse videos such as dogs running on grass,  a person walking past a building,  a woman walking on the beach, swans in a lake and more. In the first three rows of Figure \ref{fig:qualitative_2}, \ourModel reconstructs textured and complex backgrounds (e.g., grass, water, sand) without the distortions and ghosting present in OmnimatteRF and DiffuEraser (see red boxes). In all rows of Figure \ref{fig:qualitative_1}, competing methods leave visible shadow remnants, semantic drift, or object traces, while \ourModel removes foregrounds cleanly and fills in semantically appropriate content.

\begin{figure}[t!]
	\centering
\includegraphics[width=\linewidth]{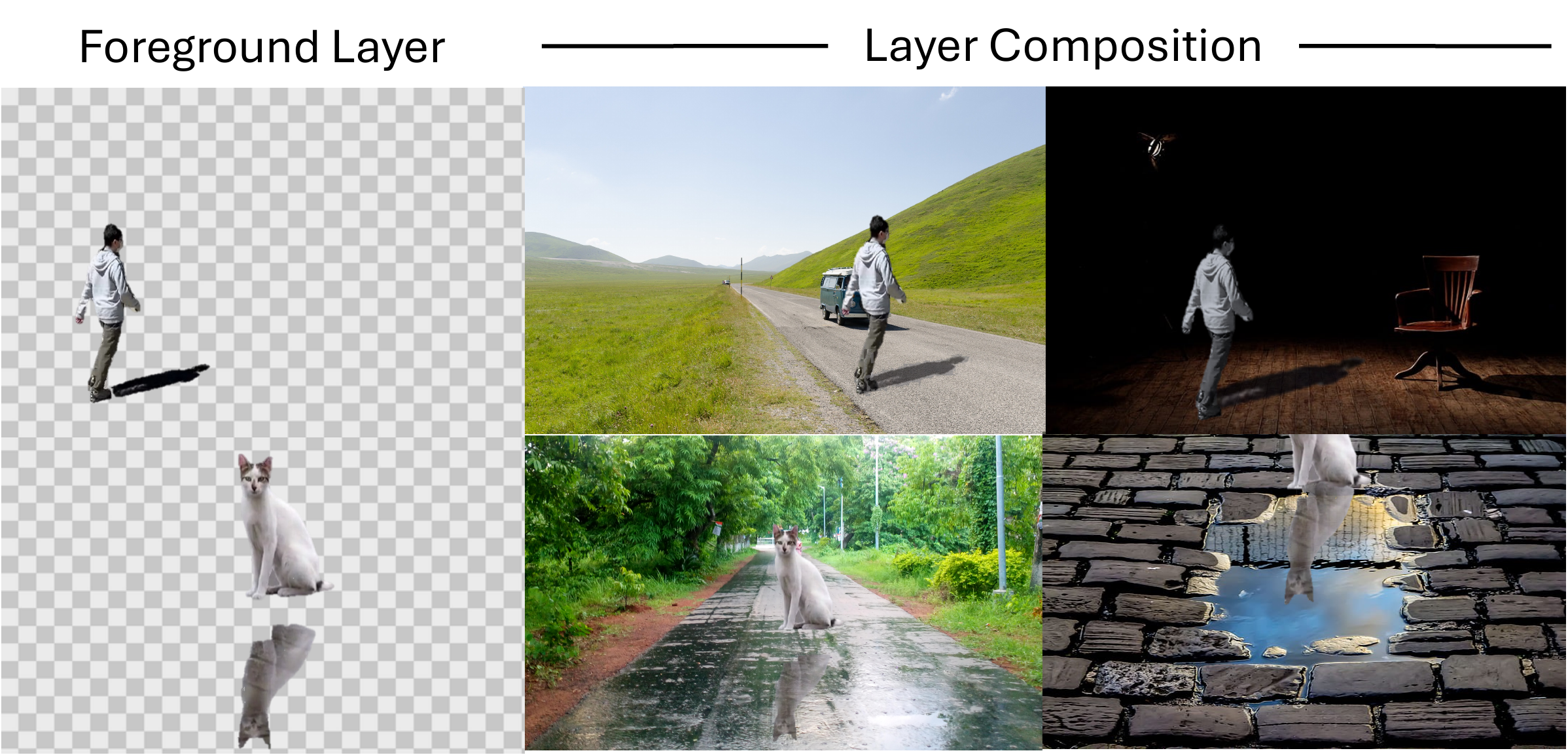}
\Description[Description]{Description}
	\caption{\textbf{Qualitative Comparison: Layer Composition.}  The extracted foreground objects, along with their shadows and reflections, are seamlessly integrated into diverse backgrounds, demonstrating the effectiveness of our approach in preserving visual coherence and realism across different scenes. See video examples in the supplemental material}
	\label{fig:qualitative_insertion}
\end{figure}

\vspace{-5pt}
\subsection{Foreground Extraction}
We aim to extract the foreground object along with its associated effects, such as shadows and reflections. \Cref{fig:qualitative_foreground} compares \ourModel for foreground extraction with OmnimatteRF \cite{omnimatteRF}. Our training-free method accurately isolates objects while preserving fine details of their interactions with the scene, such as reflections and shadows. In contrast, OmnimatteRF struggles to fully capture the associated effects. Notably, in the second and third rows, our method correctly extracts both the shadow of the cyclist and the reflection of the cat, while OmnimatteRF either distorts or omits these elements. These results demonstrate that \ourModel provides superior, training-free foreground extraction, effectively handling complex visual effects where existing methods fall short.

\subsection{Layer Composition}
The objective here is to take the extracted foreground layers and seamlessly composite them onto new background layers. Figures \ref{fig:qualitative_insertion} and \ref{fig:qualitative_3} presents the results of \ourModel, highlighting its ability to maintain object integrity, shadows, and reflections while enabling smooth re-composition across different scenes. Notably, it demonstrates how the model adaptively integrates the inserted object into the environment, such as enhancing the cat’s reflection in the clear water stain or adjusting the man's appearance to match the night lighting.


\section{Conclusion and Limitations}
\label{sec:conclusion}

In this work, we introduced \emph{OmnimatteZero}, the first training-free approach
for video omnimatte. By leveraging pre-trained video diffusion models together with
spatial and temporal attention guidance, OmnimatteZero enables real-time object
removal, foreground extraction, and seamless layer recomposition without any
additional training or test-time optimization. Extensive experiments demonstrate
that our method outperforms supervised and self-supervised baselines in both
background reconstruction and multi-object handling, while achieving state-of-the-art
runtime efficiency.

Nevertheless, several limitations remain. First, since our pipeline relies on VAE
compression and decompression, the reconstructed video may exhibit slight deviations
from the original input. Second, the quality of temporal correspondence depends on
the accuracy of TAP, which can degrade under heavy occlusions or in low-resolution
videos. Third, the fidelity of inpainting is ultimately bounded by the capabilities
of the underlying video diffusion model, which may struggle in highly complex or
unfamiliar scenes. We believe these limitations will be progressively mitigated as
future video generative models continue to improve in visual fidelity, temporal
consistency, and robustness. OmnimatteZero provides a fast, flexible, and practical
framework that can readily incorporate such advances, paving the way for scalable,
real-time video editing applications.

 \begin{acks}
We thank Yoni Kasten for his insightful input and valuable contributions to this work.
\end{acks}
\bibliographystyle{ACM-Reference-Format}
\bibliography{main}
\clearpage



\clearpage

\appendix
\newpage

\twocolumn[
    {\LARGE \bfseries OmnimatteZero: Fast Training-free Omnimatte with Pre-trained Video Diffusion Models \par}
    \vspace{1em}

    {\large 
    DVIR SAMUEL, Bar-Ilan University \& OriginAI, Israel \par
    MATAN LEVY, The Hebrew University of Jerusalem, Israel \par
    NIR DARSHAN, OriginAI, Israel \par
    GAL CHECHIK, Bar-Ilan University \& NVIDIA Research, Israel \par
    RAMI BEN-ARI, OriginAI, Israel \par
    }

    \vspace{2em}
]




\section{Implementation Details} 
We apply OmnimatteZero using two video diffusion models: LTXVideo \cite{LTXVideo} v0.9.1 and Wan2.1~\cite{wan2.1}, both running with \( T_{\text{steps}} = 30 \) denoising steps. The guidance scale was set to 0 (\ie no prompt) as we found no major effect of the prompt on the process, and all result videos were generated using the same random seed.

Quantitative and qualitative results of baselines were obtained from the respective authors.  All qualitative results in this paper are based on LTXVideo, as it shows minimal visual differences from Wan2.1. Additionally, due to space constraints, each figure displays a single frame per video for qualitative results. Full videos are available in the supplementary material.

\textit{Latent masking:} Applying a given mask video directly in latent space is challenging due to the unclear mapping from pixel space to the latent space. We start by taking the object masks (per-frame) and computing a binary video, where the frames are binary images, and pass this video as an RGB input (with duplicated channels) through the VAE encoder as well. This process produces a latent representation where high values indicate the masked object, enabling its spatio-temporal footprint identification via simple thresholding.


\section{Ablation Study}

To assess the individual contributions of our key architectural components, we perform a series of ablation studies on the Movie benchmark using the LTXVideo dataset~\cite{LTXVideo}. Quantitative results are reported in terms of PSNR.

\subsection{Effect of Temporal and Spatial Attention Guidance}

We first analyze the impact of temporal and spatial attention mechanisms on reconstruction quality:

\begin{table}[h!]
\centering
\caption{Effect of temporal and spatial attention guidance. Temporal guidance contributes most significantly, with the highest quality achieved when both cues are combined.}
\begin{tabular}{ccc}
\toprule
\textbf{Temporal} & \textbf{Spatial} & \textbf{PSNR} \\
\midrule
\xmark & \xmark & 18.5 \\
\cmark & \xmark & 28.3 \\
\xmark & \cmark & 23.2 \\
\cmark & \cmark & \textbf{35.11 (Ours) } \\
\bottomrule
\end{tabular}

\end{table}

Results indicate that temporal attention is the primary contributor to reconstruction fidelity, with a PSNR improvement of nearly 10 points over the baseline. Spatial attention offers complementary benefits. When integrated, the two mechanisms produce a substantial synergistic effect, resulting in the highest overall performance.

\subsection{Influence of Point Sampling in Track-Any-Point}

Next, we evaluate the effect of varying the sampling density of points from the object mask in the Track-Any-Point module:

\begin{table}[h!]
\centering
\caption{Effect of point sampling density. Using 60\% of object pixels is sufficient to match the maximum reconstruction quality.}
\begin{tabular}{cc}
\toprule
\textbf{Percent of Object Pixels} & \textbf{PSNR} \\
\midrule
20\%  & 22.6 \\
40\%  & 28.1 \\
60\%  & \textbf{35.11} \\
80\%  & 35.11 \\
100\% & 35.12 \\
\bottomrule
\end{tabular}

\end{table}

We observe that increasing the proportion of sampled object pixels leads to improved reconstruction quality, with performance saturating at approximately 60\% coverage. This suggests that high-fidelity results can be attained with moderately dense sampling, highlighting the efficiency of our approach in establishing accurate correspondences.

\subsection{Impact of Effect-Aware Masking}

We investigate the influence of explicitly incorporating associated object-induced visual effects—such as shadows and reflections—into the masking process:

\begin{table}[h!]
\centering
\caption{Effect of including associated visual effects (e.g., shadows, reflections) in the mask. These effects are essential for clean background reconstruction.}
\begin{tabular}{cc}
\toprule
\textbf{Effect Masking} & \textbf{PSNR} \\
\midrule
Without & 29.16 \\
With (Ours) & \textbf{35.11} \\
\bottomrule
\end{tabular}

\end{table}

\begin{figure*}[t!]
	\centering
\includegraphics[width=0.7\linewidth]{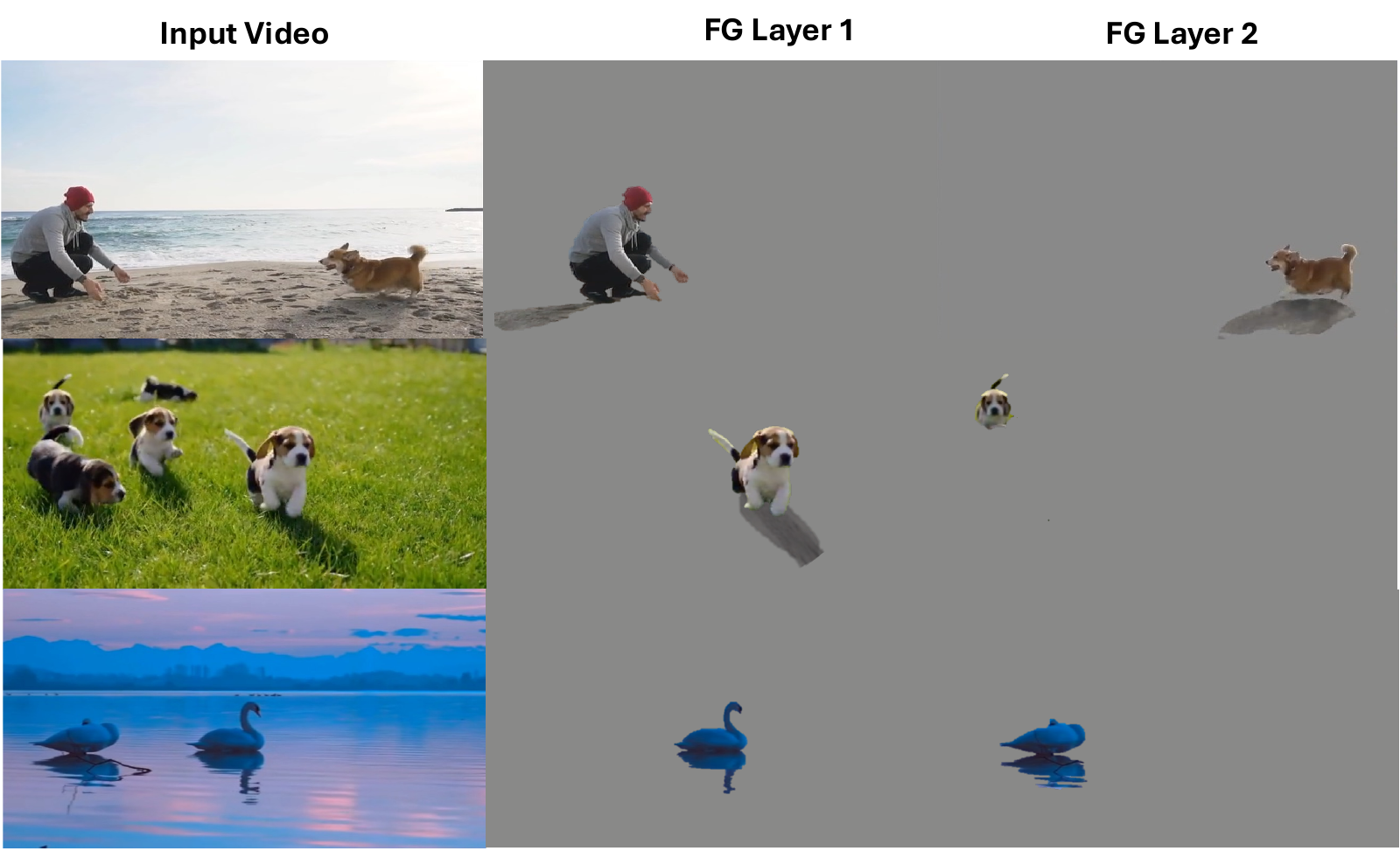}
	\Description[Description]{Description}
    \caption{Qualitative examples of \textbf{multi-object extraction}. 
    OmnimatteZero successfully separates multiple dynamic foreground objects 
    (e.g., person, dog, puppies) along with their associated effects such as 
    shadows. The results demonstrate that 
    our method scales to complex scenes with several independently moving 
    objects, preserving fine details and temporal consistency.}
	\label{fig:multi_object}
\end{figure*}

The inclusion of effect-aware masks yields a substantial improvement of +6.0 PSNR, highlighting the critical role of accurately capturing the extended influence of foreground objects on their surrounding environment. This result underscores the importance of modeling object footprints for realistic scene reconstruction.

\section{Multi-Object Extraction}
Figure~\ref{fig:multi_object} provides qualitative examples of multiple foreground 
extractions from a single video. It shows that OmnimatteZero can reliably isolate 
several dynamic objects together with their associated effects, such as shadows and 
reflections. Our 
method maintains temporal consistency and preserves fine details across all layers. 
This demonstrates the scalability of our framework to complex multi-object scenes. 
Full video results are provided in the supplementary material.

\section{Handling Object Occlusion}
We further evaluated the ability of our method to handle object occlusion, a 
particularly challenging scenario for video decomposition. To this end, we 
constructed a benchmark of 20 videos containing significant occlusions between 
objects. In these experiments, our method successfully produced coherent object 
removal and background reconstruction in 60\% of the cases. This substantially 
outperforms Generative-Omnimatte~\cite{generative-omnimatte}, which succeeded in only 
30\% of cases, and OmnimatteRF~\cite{omnimatteRF}, which achieved just 1\%. 
These results indicate that the proposed spatial and temporal attention guidance 
enables improved robustness to occlusions compared to prior approaches. Nonetheless, 
occlusion remains a difficult challenge, as heavy overlaps can degrade tracking 
accuracy and limit the available background context. We believe that future 
advancements in video generative models and more robust tracking mechanisms will 
further enhance performance in these scenarios.

\section{Comparison to Generative-Omnimatte} \label{sec:gen_omnimatte_comparison} 
Here we provide further discussion and comparison with Generative-Omnimatte~\cite{generative-omnimatte}.
Generative-Omnimatte is a supervised video diffusion method for object removal. While effective in certain cases, it requires annotated training data, per-video test-time optimization, and incurs several seconds of computation per frame. Moreover, it often struggles in scenes containing multiple objects and complex interactions. In contrast, our approach is training-free, requires no test-time optimization, and achieves real-time performance (0.04 seconds per frame on an A100 GPU). Thanks to the proposed spatial and temporal attention guidance, our method can robustly handle scenes with multiple objects while maintaining both temporal consistency and accurate background reconstruction. To further evaluate this difference, we conducted an additional comparison on 20 challenging videos, each containing more than five objects. In these settings, we attempted to remove several objects simultaneously. Our method succeeded in all cases, whereas Generative-Omnimatte failed on 90\% of the videos, leaving visible artifacts or incomplete removals. These results highlight the practical advantage of our approach in complex, multi-object scenarios, where supervised and optimization-based methods face severe limitations.
\newpage
\begin{figure*}[!h]
	\centering
\includegraphics[width=0.93\linewidth]{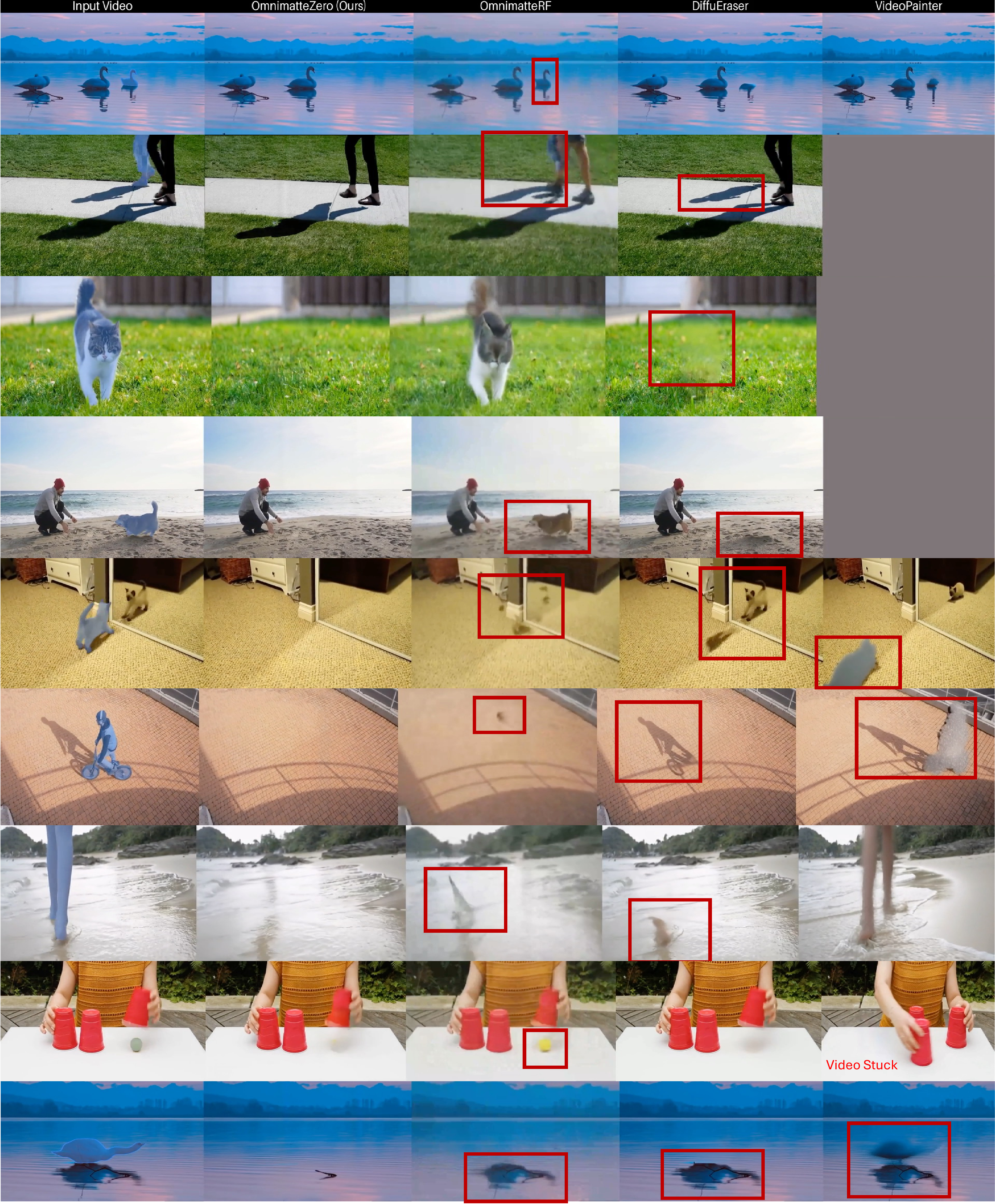}
\Description[Description]{Description}
	\caption{\textbf{Qualitative comparison with state-of-the-art video inpainting methods.} OmnimatteZero produces clean, temporally consistent background reconstructions across a diverse range of scenes, including dynamic motion, shadows, reflections, and fine textures. Compared to other methods, it avoids common artifacts such as ghosting, blur, and shadow remnants (highlighted in red boxes), successfully filling in complex backgrounds across varying temporal and spatial contexts. Gray frames indicate failure cases where the inpainting method returned an error. See the supplementary material for full videos.}
	\label{fig:qualitative_1}
\end{figure*}

\begin{figure*}[!h]
	\centering
\includegraphics[width=0.9\linewidth]{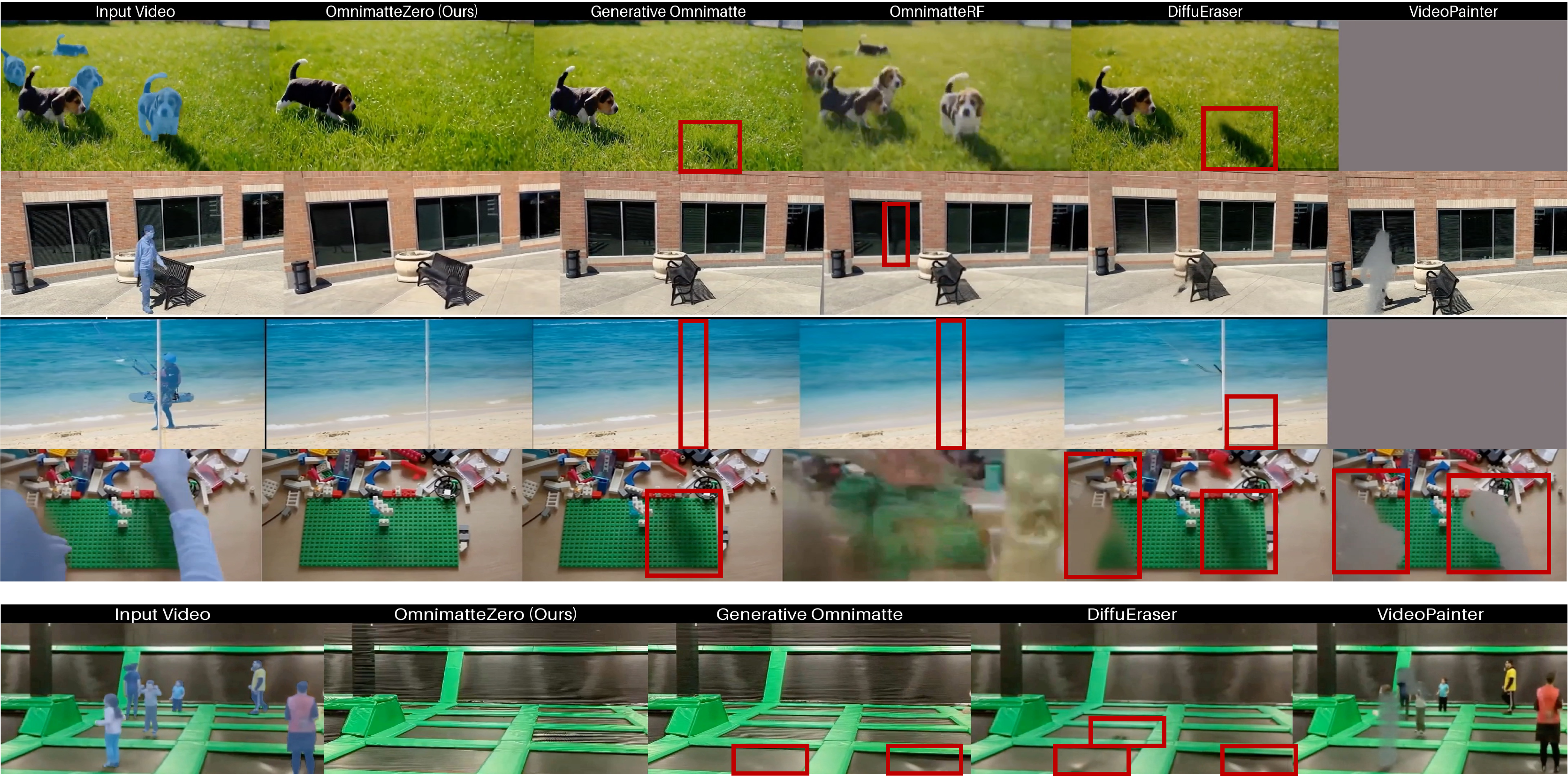}
\Description[Description]{Description}
	\caption{\textbf{Qualitative comparison with state-of-the-art video inpainting methods.}  \ourModel\ achieves clean, temporally consistent background reconstructions across diverse scenes, preserving fine textures, shadows, and reflections. Compared to prior methods, it avoids common artifacts such as ghosting, blur, and residual shadows (highlighted in red). Gray frames indicate failure cases where the method returned an error. See supplementary material for full video results.}
	\label{fig:qualitative_2}
\end{figure*}

\begin{figure*}[!h]
	\centering
\includegraphics[width=0.9\linewidth]{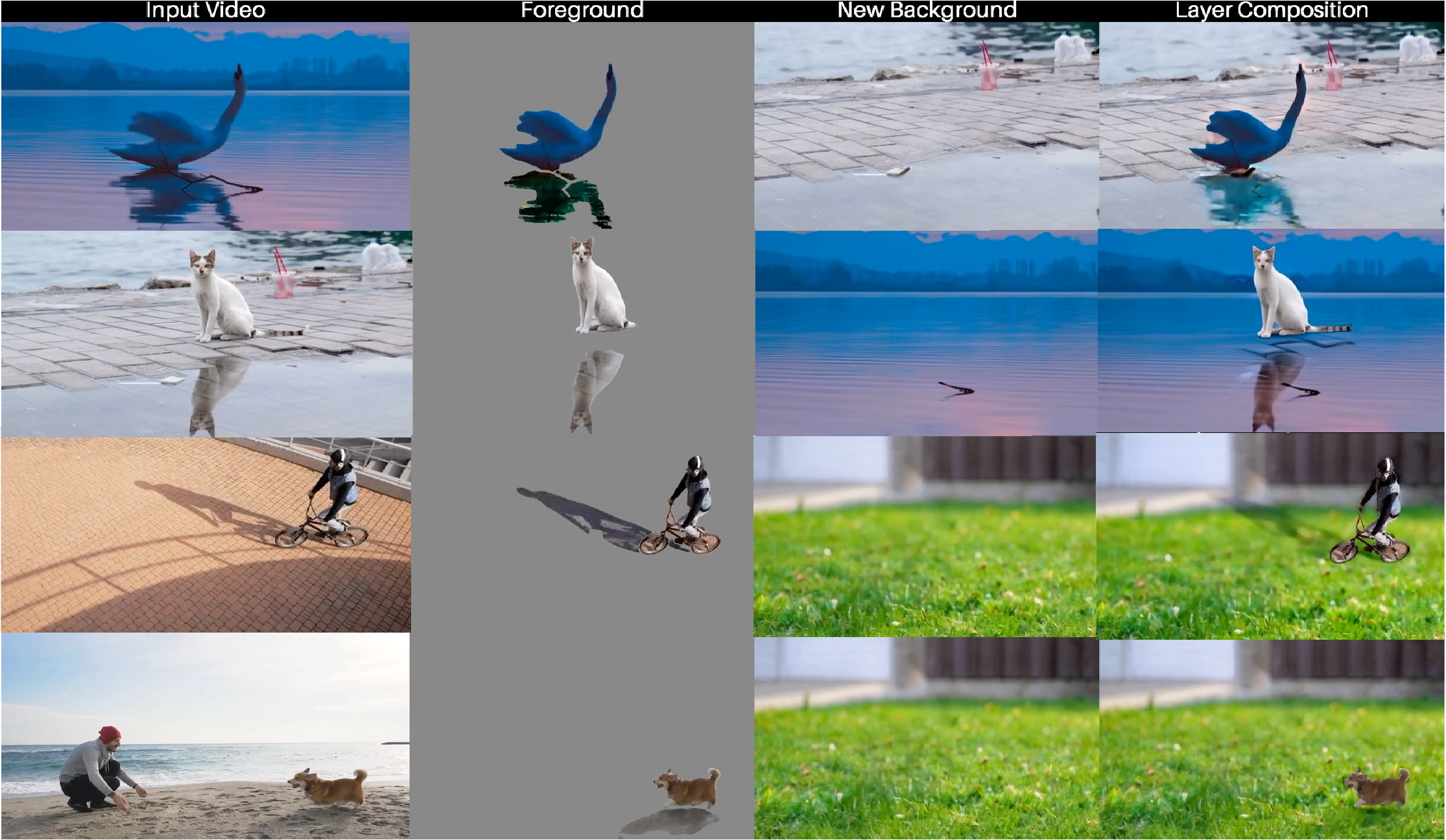}
	\Description[Description]{Description}
    \caption{\textbf{Qualitative Comparison: Layered Composition.}
From left to right: the original video frame, the extracted foreground (including shadows and reflections), the target background, and our final composite.  Across four diverse examples (a swan, a cat, a bicyclist, and a dog), our method preserves accurate shadows, reflections, and object–scene coherence, yielding visually seamless results.  See the supplementary material for full videos.
}
	\label{fig:qualitative_3}
\end{figure*}



\end{document}